\documentclass[10pt,twocolumn,letterpaper]{article}

\usepackage{cvpr}
\usepackage{times}
\usepackage{array}
\usepackage{booktabs}
\usepackage{multirow}
\usepackage{amsmath}
\usepackage{amssymb}
\usepackage{graphicx}
\usepackage{epsfig}
\usepackage{xcolor}
\usepackage{colortbl}
\usepackage[breaklinks=true,bookmarks=false]{hyperref}

\cvprfinalcopy % *** Uncomment this line for the final submission

\def\cvprPaperID{2224} % *** Enter the CVPR Paper ID here

\ifcvprfinal\fi

\begin{document}

\title{Robust Scene Text Recognition with Automatic Rectification}

\author{Baoguang Shi, Xinggang Wang, Pengyuan Lyu, Cong Yao, Xiang Bai\thanks{Corresponding author}\\
School of Electronic Information and Communications\\
Huazhong University of Science and Technology\\
{\tt\small shibaoguang@gmail.com, xbai@hust.edu.cn}}

\maketitle

\begin{abstract}
Recognizing text in natural images is a challenging task with many unsolved problems. Different from those in documents, words in natural images often possess irregular shapes, which are caused by perspective distortion, curved character placement, etc. We propose RARE (Robust text recognizer with Automatic REctification), a recognition model that is robust to irregular text. RARE is a specially-designed deep neural network, which consists of a Spatial Transformer Network (STN) and a Sequence Recognition Network (SRN). In testing, an image is firstly rectified via a predicted Thin-Plate-Spline (TPS) transformation, into a more ``readable'' image for the following SRN, which recognizes text through a sequence recognition approach. We show that the model is able to recognize several types of irregular text, including perspective text and curved text. RARE is end-to-end trainable, requiring only images and associated text labels, making it convenient to train and deploy the model in practical systems. State-of-the-art or highly-competitive performance achieved on several benchmarks well demonstrates the effectiveness of the proposed model.
\end{abstract}

\section{Introduction}
In natural scenes, text appears on various kinds of objects, \emph{e.g.} road signs, billboards, and product packaging. It carries rich and high-level semantic information that is important for image understanding. Recognizing text in images facilitates many real-world applications, such as geo-location, driverless car, and image-based machine translation. For these reasons, scene text recognition has attracted great interest from the community~\cite{NeumannM12, WangWCN12, JaderbergSVZ14}. Despite the maturity of the research on Optical Character Recognition (OCR)~\cite{Nagy00}, recognizing text in natural images, rather than scanned documents, is still challenging. Scene text images exhibit large variations in the aspects of illumination, motion blur, text font, color, \emph{etc}. Moreover, text in the wild may have irregular shape. For example, some scene text is \emph{perspective text}~\cite{PhanSTT13}, which is caused by side-view camera angles; some has curved shapes, meaning that its characters are placed along curves rather than straight lines. We call such text \emph{irregular text}, in contrast to \emph{regular text} which is horizontal and frontal.

\begin{figure}[t]
  \begin{centering}
  \includegraphics[width=1\linewidth]{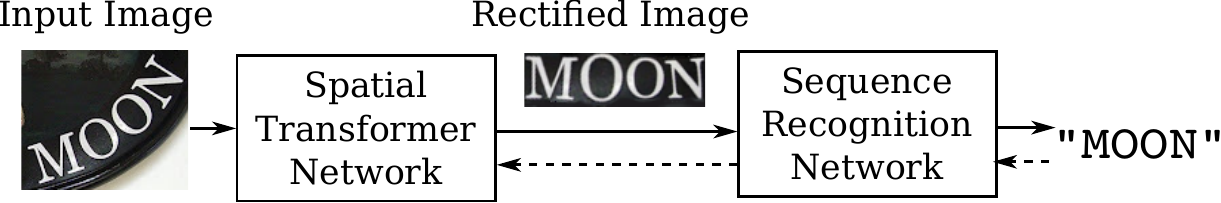}
  \end{centering}
  \caption{Schematic overview of RARE, which consists a spatial transformer network (STN) and a sequence recognition network (SRN). The STN transforms an input image to a rectified image, while the SRN recognizes text. The two networks are jointly trained by the back-propagation algorithm~\cite{LecunLYP1998}. The dashed lines represent the flows of the back-propagated gradients.}
  \label{fig:overview}
\end{figure}

Usually, a text recognizer works best when its input images contain tightly-bounded regular text. This motivates us to apply a spatial transformation prior to recognition, in order to rectify input images into ones that are more ``readable'' by recognizers. In this paper, we propose a recognition method that is robust to irregular text. Specifically, we construct a deep neural network that combines a Spatial Transformer Network~\cite{JaderbergSZK15} (STN) and a Sequence Recognition Network (SRN). An overview of the model is given in Fig.~\ref{fig:overview}.

In the STN, an input image is spatially transformed into a \emph{rectified image}. Ideally, the STN produces an image that contains regular text, which is a more appropriate input for the SRN than the original one. The transformation is a thin-plate-spline~\cite{Bookstein89} (TPS) transformation, whose nonlinearity allows us to rectify various types of irregular text, including perspective and curved text. The TPS transformation is configured by a set of \emph{fiducial points}, whose coordinates are regressed by a convolutional neural network.

In an image that contains regular text, characters are arranged along a horizontal line. It bares some resemblance to a sequential signal. Motivated by this, for the SRN we construct an \emph{attention-based} model~\cite{BahdanauCB14} that recognizes text in a sequence recognition approach. The SRN consists of an encoder and a decoder. Given an input image, the encoder generates a sequential feature representation, which is a sequence of feature vectors. The decoder recurrently generates a character sequence conditioning on the input sequence, by decoding the relevant contents which are determined by its attention mechanism at each step.

We show that, with proper initialization, the whole model can be trained end-to-end. Consequently, for the STN, we do not need to label any geometric ground truth, \emph{i.e.} the positions of the TPS fiducial points, but let its training be supervised by the error differentials back-propagated by the SRN. In practice, the training eventually makes the STN tend to produce images that contain regular text, which are desirable inputs for the SRN.

The contributions of this paper are three-fold: First, we propose a novel scene text recognition method that is robust to irregular text. Second, our model extends the STN framework~\cite{JaderbergSZK15} with an attention-based model. The original STN is only tested on plain convolutional neural networks. Third, our model adopts a convolutional-recurrent structure in the encoder of the SRN, thus is a novel variant of the attention-based model~\cite{BahdanauCB14}.

\section{Related Work}

In recent years, a rich body of literature concerning scene text recognition has been published. Comprehensive surveys have been given in \cite{YeD15, ZhuYB16}. Among the traditional methods, many adopt bottom-up approaches, where individual characters are firstly detected using sliding window~\cite{WangB10, WangBB11}, connected components~\cite{NeumannM12}, or Hough voting~\cite{YaoBSL14}. Following that, the detected characters are integrated into words by means of dynamic programming, lexicon search~\cite{WangBB11}, \emph{etc.}. Other work adopts top-down approaches, where text is directly recognized from entire input images, rather than detecting and recognizing individual characters. For example, Alm\'azan \emph{et al.}~\cite{AlmazanGFV14} propose to predict label embedding vectors from input images. Jaderberg \emph{et al.}~\cite{JaderbergSVZ14a} address text recognition with a 90k-class convolutional neural network, where each class corresponds to an English word. In~\cite{JaderbergSVZ14b}, a CNN with a structured output layer is constructed for unconstrained text recognition. Some recent work models the problem as a sequence recognition problem, where text is represented by character sequence. Su and Lu~\cite{SuL14} extract sequential image representation, which is a sequence of HOG~\cite{DalalT05} descriptors, and predict the corresponding character sequence with a recurrent neural network (RNN). Shi \emph{et al.}~\cite{ShiBY15} propose an end-to-end sequence recognition network which combines CNN and RNN. Our method also adopts the sequence prediction scheme, but we further take the problem of irregular text into account.

Although being common in the tasks of scene text detection and recognition, the issue of irregular text is relatively less addressed in explicit ways. Yao \emph{et al.}~\cite{YaoBLMT12} firstly propose the multi-oriented text detection problem, and deal with it by carefully designing rotation-invariant region descriptors. Zhang \emph{et al.}~\cite{ZhangGLM12} propose a character rectification method that leverages the low-rank structures of text. Phan \emph{et al.} propose to explicitly rectify perspective distortions via SIFT~\cite{Lowe04} descriptor matching. The above-mentioned work brings insightful ideas into this issue. However, most methods deal with only one type of irregular text with specifically designed schemes. Our method rectifies several types of irregular text in a unified way. Moreover, it does not require extra annotations for the rectification process, since the STN is supervised by the SRN during training.

\section{Proposed Model}

In this section we formulate our model. Overall, the model takes an input image $I$ and outputs a sequence $\mathbf{l} = ( l_{1},\dots,l_{T} )$, where $l_{t}$ is the $t$-th character, $T$ is the variable string length.

\subsection{Spatial Transformer Network}

\begin{figure}[ht]
  \begin{centering}
  \includegraphics[width=1\linewidth]{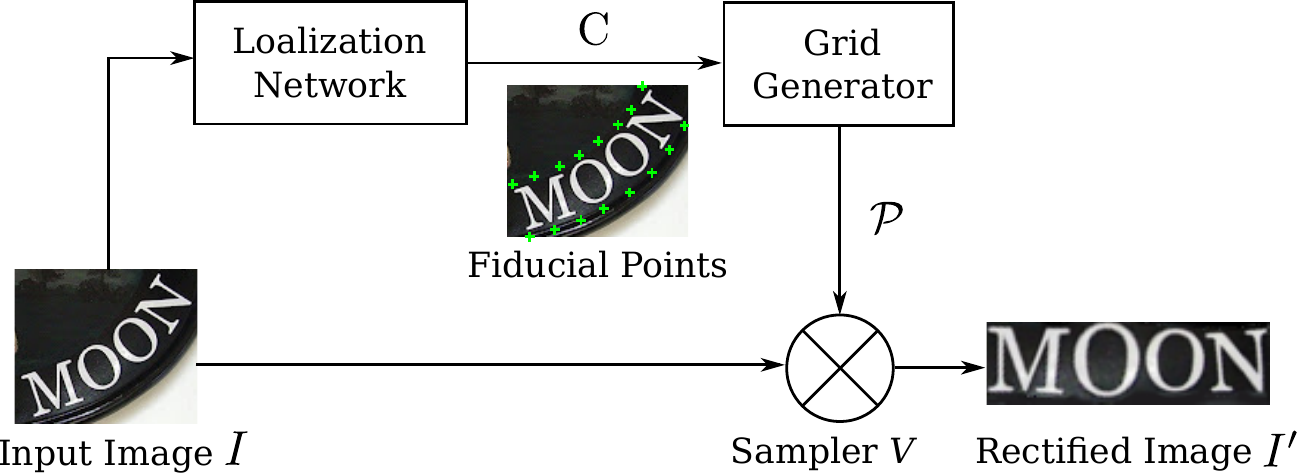}
  \end{centering}
  \caption{Structure of the STN. The localization network localizes a set of fiducial points $\mathbf{C}$, with which the grid generator generates a sampling grid ${\cal P}$. The sampler produces a rectified image $I'$, given $I$ and ${\cal P}$.}
  \label{fig:STN}
\end{figure}

The STN transforms an input image $I$ to a rectified image $I'$ with a predicted TPS transformation. It follows the framework proposed in~\cite{JaderbergSZK15}. As illustrated in Fig.~\ref{fig:STN}, it first predicts a set of fiducial points via its localization network. Then, inside the grid generator, it calculates the TPS transformation parameters from the fiducial points, and generates a sampling grid on $I$. The sampler takes both the grid and the input image, it produces a rectified image $I'$ by sampling on the grid points.

A distinctive property of STN is that its sampler is differentiable. Therefore, once we have a differentiable localization network and a differentiable grid generator, the STN can back-propagate error differentials and gets trained.

\subsubsection{Localization Network}

The localization network localizes $K$ fiducial points by directly regressing their $x,y$-coordinates. Here, constant $K$ is an even number. The coordinates are denoted by $\mathbf{C} = \left[ \mathbf{c}_{1}, \dots, \mathbf{c}_{K} \right] \in \Re^{2 \times K}$, whose $k$-th column $\mathbf{c}_{k}=\left[x_{k},y_{k}\right]^{\intercal}$ contains the coordinates of the $k$-th fiducial point. We use a normalized coordinate system whose origin is the image center, so that $x_k, y_k$ are within the range of $[-1, 1]$.

We use a convolutional neural network (CNN) for the regression.
% , which is formulated as
% \begin{equation}
%   \mathbf{C} = \mathrm{CNN}(I,\boldsymbol{\theta}_{\mathrm{loc}}).
%   \label{eq:loc-net}
% \end{equation}
% Here, $\boldsymbol{\theta}_{\mathrm{loc}}$ is the parameters of the CNN.
Similar to the conventional structures~\cite{SimonyanZ14a,KrizhevskySH12}, the CNN contains convolutional layers, pooling layers and fully-connected layers. However, we use it for regression instead of classification. For the output layer, which is the last fully-connected layer, we set the number of output nodes to $2K$ and the activation function to $\tanh(\cdot)$, so that its output vectors have values that are within the range of $\left( -1, 1 \right)$. Last, the output vector is reshaped into $\mathbf{C}$.

The network localizes fiducial points based on global image contexts. It is expected to capture the overall text shape of an input image, and localizes fiducial points accordingly. It should be emphasized that we do not annotate coordinates of fiducial points for any sample. Instead, the training of the localization network is completely supervised by the gradients propagated by the other parts of the STN, following the back-propagation algorithm~\cite{LecunLYP1998}.

\subsubsection{Grid Generator}

\begin{figure}[ht]
  \begin{centering}
  \includegraphics[width=0.85\linewidth]{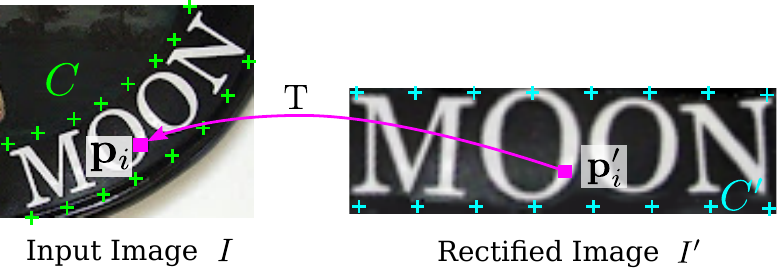}
  \par\end{centering}
  \vspace{0.1cm}
  \caption{Fiducial points and the TPS transformation. Green markers on the left image are the fiducial points $\mathbf{C}$. Cyan markers on the right image are the base fiducial points $\mathbf{C}'$. The transformation $\mathbf{T}$ is represented by the pink arrow. For a point $(x_{i}',y_{i}')$ on $I'$, the transformation $\mathbf{T}$ finds the corresponding point $(x_{i},y_{i})$ on $I$.}
  \label{fig:fiducial-trans}
\end{figure}

The grid generator estimates the TPS transformation parameters, and generates a sampling grid. We first define another set of fiducial points, called the \emph{base fiducial points}, denoted by $\mathbf{C}'=\left[\mathbf{c}_{1}',\dots,\mathbf{c}_{K}'\right]\in\Re^{2\times K}$. As illustrated in Fig.~\ref{fig:fiducial-trans}, the base fiducial points are evenly distributed along the top and bottom edge of a rectified image $I'$. Since $K$ is a constant and the coordinate system is normalized, $\mathbf{C}'$ is always a constant.

The parameters of the TPS transformation is represented by a matrix $\mathbf{T}\in\Re^{2\times(K+3)}$, which is computed by
\begin{equation}
  \mathbf{T}=\left(\boldsymbol{\Delta}_{\mathbf{C}'}^{-1}\left[\begin{array}{c}
    \mathbf{C}^{\intercal}\\
    \mathbf{0}^{3\times2}
    \end{array}
  \right]\right)^{\intercal},
  \label{eq:tps-solve}
\end{equation}
where $\boldsymbol{\Delta}_{\mathbf{C}'}\in\Re^{(K+3)\times(K+3)}$ is a matrix determined only by $\mathbf{C}'$, thus also a constant:
\[
\boldsymbol{\Delta}_{\mathbf{C}'}=\begin{bmatrix}\mathbf{1}^{K\times1} & \mathbf{C}'^{\intercal} & \mathbf{R}\\
\mathbf{0} & \mathbf{0} & \mathbf{1}^{1\times K}\\
\mathbf{0} & \mathbf{0} & \mathbf{C}'
\end{bmatrix},
\]
where the element on the $i$-th row and $j$-th column of $\mathbf{R}$ is $r_{i,j}=d_{i,j}^{2} \ln d_{i,j}^{2}$, $d_{i,j}$ is the euclidean distance between $\mathbf{c}'_{i}$ and $\mathbf{c}'_{j}$.

The grid of pixels on a rectified image $I'$ is denoted by ${\cal P}'=\left\{ \mathbf{p}'_{i} \right\} _{i=1,\dots,N}$, where $\mathbf{p}_{i}'=\left[x_{i}',y_{i}'\right]^{\intercal}$ is the x,y-coordinates of the $i$-th pixel, $N$ is the number of pixels. As illustrated in Fig.~\ref{fig:fiducial-trans}, for every point $\mathbf{p}'_{i}$ on $I'$, we find the corresponding point $\mathbf{p}_{i}=\left[x_{i},y_{i}\right]^{\intercal}$ on $I$, by applying the transformation:
\begin{align}
  r_{i,k}' & = d_{i,k}^{2}\ln d_{i,k}^{2}\\
  \hat{\mathbf{p}}'_{i} & =\left[1,x_{i}',y_{i}',r_{i,1}',\dots,r_{i,K}'\right]^{\intercal}\label{eq:tps-coord}\\
  \mathbf{p}_{i} & = \mathbf{T}\hat{\mathbf{p}}'_{i}\label{eq:tps-trans},
\end{align}
where $d_{i,k}$ is the euclidean distance between $\mathbf{p}_{i}'$ and the $k$-th base fiducial point $\mathbf{c}'_{k}$.

By iterating over all points in ${\cal P}'$, we generate a grid ${\cal P}=\left\{ \mathbf{p}{}_{i}\right\} _{i=1,\dots,N}$ on the input image $I$. The grid generator can back-propagate gradients, since its two matrix multiplications,  Eq.~\ref{eq:tps-solve} and Eq.~\ref{eq:tps-trans}, are both differentiable.

\subsubsection{Sampler}

Lastly, in the sampler, the pixel value of $\mathbf{p}_{i}'$ is bilinearly interpolated from the pixels near $\mathbf{p}_{i}$ on the input image. By setting all pixel values, we get the rectified image $I'$:
\begin{equation}
  I'=V({\cal P},I)
  \label{eq:tps-sampler},
\end{equation}
where $V$ represents the bilinear sampler~\cite{JaderbergSZK15}, which is also a differentiable module.

\begin{figure}[t]
  \begin{centering}
  \includegraphics[width=0.8\linewidth]{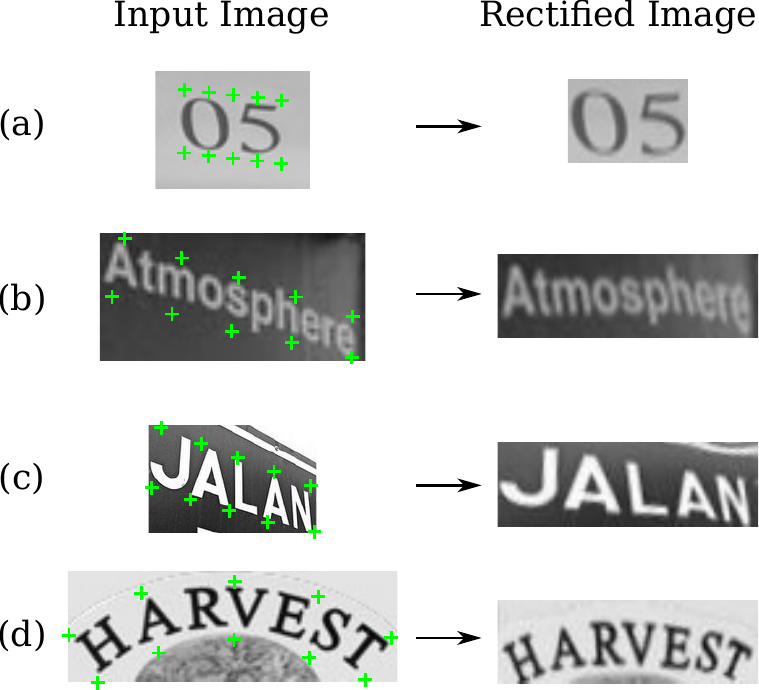}
  \par\end{centering}
  \vspace{0.2cm}
  \caption{The STN rectifies images that contain several types of irregular text. Green markers are the predicted fiducial points on the input images. The STN can deal with several types of irregular text, including (a) loosely-bounded text; (b) multi-oriented text; (c) perspective text; (d) curved text.}
  \label{fig:tps-example}
\end{figure}

The flexibility of the TPS transformation allows us to transform irregular text images into rectified images that contain regular text. In Fig.~\ref{fig:tps-example}, we show some common types of irregular text, including a) loosely-bounded text, which resulted by imperfect text detection; b) multi-oriented text, caused by non-horizontal camera views; c) perspective text, caused by side-view camera angles; d) curved text, a commonly seen artistic style. The STN is able to rectify images that contain these types of irregular text, making them more readable for the following recognizer.

\subsection{Sequence Recognition Network}

Since target words are inherently sequences of characters, we model the recognition problem as a sequence recognition problem, and address it with a sequence recognition network. The input to the SRN is a rectified image $I'$, which ideally contains a word that is written horizontally from left to right. We extract a sequential representation from $I'$, and recognize a word from it.

In our model, the SRN is an \emph{attention-based model}~\cite{BahdanauCB14, ChorowskiBSCB15}, which directly recognizes a sequence from an input image. The SRN consists of an encoder and a decoder. The encoder extracts a sequential representation from the input image $I'$. The decoder recurrently generates a sequence conditioned on the sequential representation, by decoding the relevant contents it attends to at each step.

\subsubsection{Encoder: Convolutional-Recurrent Network}

A na\"ive approach for extracting a sequential representation for $I'$ is to take local image patches from left to right, and describe each of them with a CNN. However, this approach does not share the computation among overlapping patches, thus inefficient. Besides, the spatial dependencies between the patches are not exploited and leveraged. Instead, following~\cite{ShiBY15}, we build a network that combines convolutional layers and recurrent networks. The network extracts a sequence of feature vectors, given an input image of arbitrary size.

\begin{figure}[t]
  \begin{centering}
  \includegraphics[width=0.8\linewidth]{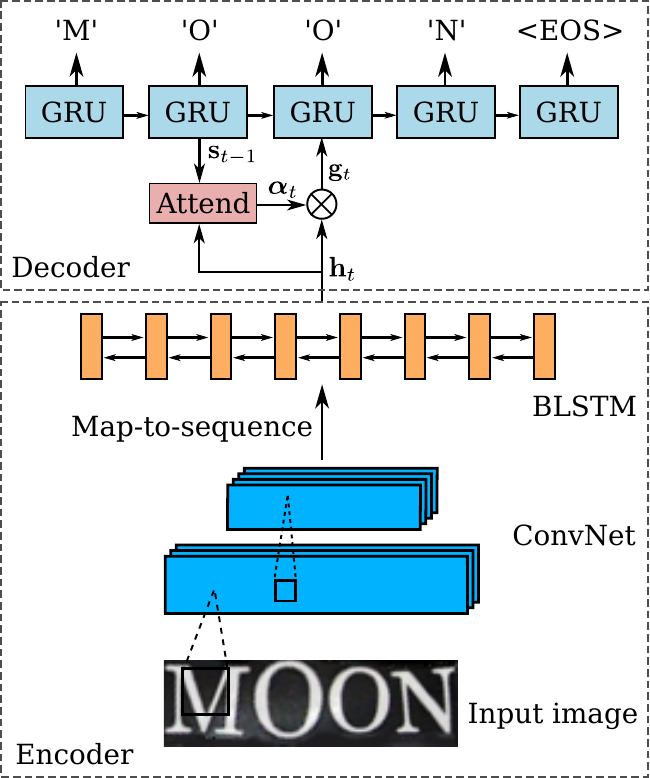}
  \par\end{centering}
  \vspace{0.2cm}
  \caption{Structure of the SRN, which consists of an encoder and a decoder. The encoder uses several convolution layers (ConvNet) and a two-layer BLSTM network to extract a sequential representation ($\mathbf{h}$) for the input image. The decoder generates a character sequence (including the EOS token) conditioned on $\mathbf{h}$.}
  \label{fig:recog-structure}
\end{figure}

As illustrated in Fig.~\ref{fig:recog-structure}, at the bottom of the encoder is several convolutional layers. They produce feature maps that are robust and high-level descriptions of an input image. Suppose the feature maps have the size $D_{\mathrm{conv}} \times H_{\mathrm{conv}} \times W_{\mathrm{conv}}$, where $D_{\mathrm{conv}}$ is the depth, and $H_{\mathrm{conv}}$, $W_{\mathrm{conv}}$ are the height and width respectively. The next operation is to convert the maps into a sequence of $W_{\mathrm{conv}}$ vectors, each has $D_{\mathrm{conv}} W_{\mathrm{conv}}$ dimensions. Specifically, the ``map-to-sequence'' operation takes out the columns of the maps in the left-to-right order, and flattens them into vectors. According to the translation invariance property of CNN, each vector corresponds to a local image region, \emph{i.e.} \emph{receptive field}, and is a descriptor for that region.

Restricted by the sizes of the receptive fields, the feature sequence leverages limited image contexts. We further apply a two-layer Bidirectional Long-Short Term Memory (BLSTM)~\cite{HochreiterS97,GravesMH13} network to the sequence, in order to model the long-term dependencies within the sequence. The BLSTM is a recurrent network that can analyze the dependencies within a sequence in both directions, it outputs another sequence which has the same length as the input one. The output sequence is $\mathbf{h} = ( \mathbf{h}_{1},\dots,\mathbf{h}_{L} )$, where $L=W_{\mathrm{conv}}$.

\subsubsection{Decoder: Recurrent Character Generator}

The decoder recurrently generates a sequence of characters, conditioned on the sequence produced by the encoder. It is a recurrent neural network with the attention structure proposed in~\cite{BahdanauCB14, ChorowskiBSCB15}. In the recurrency part, we adopt the Gated Recurrent Unit (GRU)~\cite{ChoMBB14} as the cell.

The generation is a $T$-step process, at step $t$, the decoder computes a vector of attention weights $\boldsymbol{\alpha}_{t} \in \Re^{L}$ via the attention process described in~\cite{ChorowskiBSCB15}:
\begin{equation}
  \boldsymbol{\alpha}_{t} = \mathrm{Attend} (\mathbf{s}_{t-1}, \boldsymbol{\alpha}_{t-1}, \mathbf{h}),
\end{equation}
where $\mathbf{s}_{t-1}$ is the state variable of the GRU cell at the last step. For $t=1$, both $\mathbf{s}_{0}$ and $\boldsymbol{\alpha}_{0}$ are zero vectors. Then, a \emph{glimpse} $\mathbf{g}_{t}$ is computed by linearly combining the vectors in $\mathbf{h}$: $\mathbf{g}_{t} = \sum_{i=1}^{L} \alpha_{ti} \mathbf{h}_{i}$. Since $\boldsymbol{\alpha}_{t}$ has non-negative values that sum to one, it effectively controls where the decoder focuses on.

The state $\mathbf{s}_{t-1}$ is updated via the recurrent process of GRU~\cite{ChoMBB14, ChorowskiBSCB15}:
\begin{equation}
  \mathbf{s}_{t} = \mathrm{GRU}(l_{t-1}, \mathbf{g}_{t}, \mathbf{s}_{t-1}),
\end{equation}
where $l_{t-1}$ is the $(t-1)$-th ground-truth label in training, while in testing, it is the label predicted in the previous step, \emph{i.e.} $\hat{l}_{t-1}$.

The probability distribution over the label space is estimated by:
\begin{equation}
  \hat{\mathbf{y}}_{t} = \mathrm{softmax}(\mathbf{W}^{\intercal}\mathbf{s}_{t}).
  \label{eq:gru-prob-dist}
\end{equation}
Following that, a character $\hat{l}_{t}$ is predicted by taking the class with the highest probability. The label space includes all English alphanumeric characters, plus a special ``end-of-sequence'' (EOS) token, which ends the generation process.

The SRN directly maps a input sequence to another sequence. Both input and output sequences may have arbitrary lengths. It can be trained with only word images and associated text.

\subsection{Model Training}

We denote the training set by ${\cal X}=\{ ( I^{(i)}, \mathbf{l}^{(i)} ) \}_{i=1\dots N}$. To train the model, we minimize the negative log-likelihood over ${\cal X}$:
\begin{equation}
  {\cal L} = \sum_{i=1}^{N} \log \prod_{t=1}^{|\mathbf{l}^{(i)}|} p(l_{t}^{(i)} | I^{(i)};\boldsymbol{\theta}),
\end{equation}
where the probability $p(\cdot)$ is computed by Eq.~\ref{eq:gru-prob-dist}, $\boldsymbol{\theta}$ is the parameters of both STN and SRN. The optimization algorithm is the ADADELTA~\cite{Matthew12ADADELTA}, which we find fast in convergence speed.

\begin{figure}[h]
  \begin{centering}
  \includegraphics[width=0.85\linewidth]{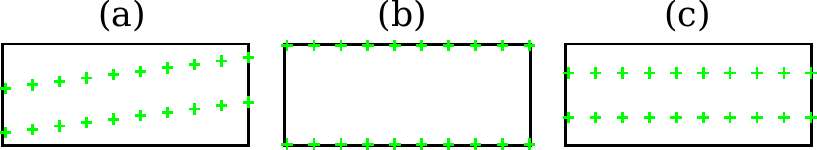}
  \par\end{centering}
  \vspace{0.2cm}
  \caption{Some initialization patterns for the fiducial points.}
  \label{fig:init-patterns}
\end{figure}

The model parameters are randomly initialized, except the localization network, whose output fully-connected layer is initialized by setting weights to zero. The initial biases are set to such values that yield the fiducial points pattern displayed in Fig.~\ref{fig:init-patterns}.a. Empirically, we also find that the patterns displayed Fig.~\ref{fig:init-patterns}.b and Fig.~\ref{fig:init-patterns}.c yield relatively poorer performance. Randomly initializing the localization network results in failure of convergence during training.

\subsection{Recognizing With a Lexicon} \label{sec:recog-lexicon}

When a test image is associated with a lexicon, \emph{i.e.} a set of words for selection, the recognition process is to pick the word with the highest posterior conditional probability:
\begin{equation}
  \mathbf{l}^{*} = \arg \max_{\mathbf{l}} \log \prod_{t=1}^{|\mathbf{l}|} p(l_{t} | I;\boldsymbol{\theta}).
  \label{fig:recog-lexicon}
\end{equation}

However, on very large lexicons, \emph{e.g.} the Hunspell~\cite{Hunspell} which contains more than 50k words, computing Eq.~\ref{fig:recog-lexicon} is time consuming, as it requires iterating over all lexicon words. We adopt an efficient approximate search scheme on large lexicons. The motivation is that computation can be shared among words that share the same prefix.

\begin{figure}[t]
  \begin{centering}
  \includegraphics[width=0.7\linewidth]{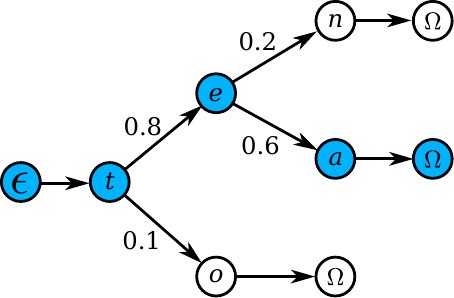}
  \par\end{centering}
  \vspace{0.2cm}
  \caption{A prefix tree of three words: ``ten'', ``tea'', and ``to''. $\epsilon$ and $\Omega$ are the tree root and the EOS token respectively. The recognition starts from the tree root. At each step the posterior probabilities of all child nodes are computed. The child node with the highest probability is selected as the next node. The process iterates until a leaf node is reached. Numbers on the edges are the posterior probabilities. Blue nodes are the selected nodes. In this case, the predicted word is ``tea''.}
  \label{fig:prefix-tree}
\end{figure}

We first construct a prefix tree over a given lexicon. As illustrated in Fig.~\ref{fig:prefix-tree}, each node of the tree is a character label. Nodes on a path from the root to a leaf forms a word (including the EOS). In testing, we start from the root node, every time the model outputs a distribution $\hat{\mathbf{y}}_{t}$, the child node with the highest posterior probability is selected as the next node to move to. The process repeats until a leaf node is reached, and a word is found on the path from the root to that leaf. Since the tree depth is at most the length of the longest word in the lexicon, this search process takes much less computation than the precise search.

Recognition performance could be further improved by incorporating beam search. A list of nodes is maintained, and the above search process is repeated on each of them. After each step, the list is updated to store the nodes with top-$B$ accumulated log-likelihoods, where $B$ is the beam width. Larger beam width usually results in better performance, but lower search speed.

\begin{table*}[t]
  \caption{Recognition accuracies on general recognition benchmarks. The titles ``50'', ``1k'' and ``50k'' are lexicon sizes. The ``Full'' lexicon contains all per-image lexicon words. ``None'' means recognition without a lexicon.}
  \footnotesize
  \vspace{0.2cm}
  \begin{centering}
  \begin{tabular}{|l||c|c|c||c|c||c|c|c|c||c|}
    \hline
    \multirow{2}{*}{\textbf{Method}} & \multicolumn{3}{c||}{\textbf{IIIT5K}} & \multicolumn{2}{c||}{\textbf{SVT}} & \multicolumn{4}{c||}{\textbf{IC03}} & \textbf{IC13} \\
    \cline{2-11}
    & \textbf{50} & \textbf{1k} & \textbf{None} & \textbf{50} & \textbf{None} & \textbf{50} & \textbf{Full} & \textbf{50k} & \textbf{None} & \textbf{None} \\
    \hline
    ABBYY \cite{WangBB11} & 24.3 & - & - & 35.0 & - & 56.0 & 55.0 & - & - & - \\
    Wang \emph{et al.} \cite{WangBB11} & - & - & - & 57.0 & - & 76.0 & 62.0 & - & - & - \\
    Mishra \emph{et al.} \cite{MishraAJ12} & 64.1 & 57.5 & - & 73.2 & - & 81.8 & 67.8 & - & - & - \\
    Wang \emph{et al.} \cite{WangWCN12} & - & - & - & 70.0 & - & 90.0 & 84.0 & - & - & - \\
    Goel \emph{et al.} \cite{GoelMAJ13} & - & - & - & 77.3 & - & 89.7 & - & - & - & - \\
    Bissacco \emph{et al.} \cite{BissaccoCNN13} & - & - & - & 90.4 & 78.0 & - & - & - & - & 87.6 \\
    Alsharif and Pineau \cite{AlsharifP13} & - & - & - & 74.3 & - & 93.1 & 88.6 & 85.1 & - & - \\
    Almaz\'an \emph{et al.} \cite{AlmazanGFV14} & 91.2 & 82.1 & - & 89.2 & - & - & - & - & - & - \\
    Yao \emph{et al.} \cite{YaoBSL14} & 80.2 & 69.3 & - & 75.9 & - & 88.5 & 80.3 & - & - & - \\
    Rodríguez-Serrano \emph{et al.} \cite{Rodriguez-Serrano15} & 76.1 & 57.4 & - & 70.0 & - & - & - & - & - & - \\
    Jaderberg \emph{et al.} \cite{JaderbergVZ14} & - & - & - & 86.1 & - & 96.2 & 91.5 & - & - & - \\
    Su and Lu \cite{SuL14} & - & - & - & 83.0 & - & 92.0 & 82.0 & - & - & - \\
    Gordo \cite{Gordo14} & 93.3 & 86.6 & - & 91.8 & - & - & - & - & - & - \\
    Jaderberg \emph{et al.} \cite{JaderbergSVZ14a} & 97.1 & 92.7 & - & 95.4 & 80.7 & \textbf{98.7} & \textbf{98.6} & 93.3 & \textbf{93.1} & \textbf{90.8} \\
    Jaderberg \emph{et al.} \cite{JaderbergSVZ14b} & 95.5 & 89.6 & - & 93.2 & 71.7 & 97.8 & 97.0 & 93.4 & 89.6 & 81.8 \\
    Shi \emph{et al.}~\cite{ShiBY15} & \textbf{97.6} & \textbf{94.4} & 78.2 & \textbf{96.4} & 80.8 & \textbf{98.7} & 97.6 & \textbf{95.5} & 89.4 & 86.7 \\
    \hline
    \textbf{RARE} & 96.2 & 93.8 & \textbf{81.9} & 95.5 & \textbf{81.9} & 98.3 & 96.2 & 94.8 & 90.1 & 88.6 \\
    \textbf{RARE (SRN only)} & 96.5 & 92.8 & 79.7 & 96.1 & 81.5 & 97.8 & 96.4 & 93.7 & 88.7 & 87.5 \\
    \hline
  \end{tabular}
  \par\end{centering}
  \label{tbl:general-datasets}
\end{table*}

\section{Experiments}

In this section we evaluate our model on a number of standard scene text recognition benchmarks, paying special attention to recognition performance on irregular text. First we evaluate our model on some general recognition benchmarks, which mainly consist of regular text, but irregular text also exists. Next, we perform evaluations on benchmarks that are specially designed for irregular text recognition. For all benchmarks, performance is measured by word accuracy.

\subsection{Implementation Details}

\noindent \textbf{Spatial Transformer Network}\hspace{0.5em}
The localization network of STN has 4 convolution layers, each followed by a $2\times2$ max-pooling layer. The filter size, padding size and stride are 3, 1, 1 respectively, for all convolutional layers. The number of filters are respectively 64, 128, 256 and 512. Following the convolutional and the max-pooling layers is two fully-connected layers with 1024 hidden units. We set the number of fiducial points to $K=20$, meaning that the localization network outputs a 40-dimensional vector. Activation functions for all layers are the ReLU~\cite{NairH10}, except the output layer which uses $\tanh(\cdot)$.

\noindent \textbf{Sequence Recognition Network}\hspace{0.5em}
In the SRN, the encoder has 7 convolutional layers, whose \{filter size, number of filters, stride, padding size\} are respectively \{3,64,1,1\}, \{3,128,1,1\}, \{3,256,1,1\}, \{3,256,1,1,\}, \{3,512,1,1\}, \{3,512,1,1\}, and \{2,512,1,0\}. The 1st, 2nd, 4th, 6th convolutional layers are each followed by a $2\times2$ max-pooling layer. On the top of the convolutional layers is a two-layer BLSTM network, each LSTM has 256 hidden units. For the decoder, we use a GRU cell that has 256 memory blocks and 37 output units (26 letters, 10 digits, and 1 EOS token).

\noindent \textbf{Model Training}\hspace{0.5em}
Our model is trained on the 8-million synthetic samples released by Jaderberg \emph{et al.}~\cite{JaderbergSVZ14}. No extra data is used. The batch size is set to 64 in training. Following~\cite{JaderbergSVZ14a, JaderbergSVZ14b}, images are resized to $100 \times 32$ in both training and testing. The output size of the STN is also $100 \times 32$. Our model processes $\sim$160 samples per second during training, and converges in 2 days after $\sim$3 epochs over the training dataset.

\noindent \textbf{Implementation}\hspace{0.5em}
We implement our model under the Torch7 framework~\cite{Collobert11}. Most parts of the model are GPU-accelerated. All our experiments are carried out on a workstation which has one Intel Xeon(R) E5-2620 2.40GHz CPU, an NVIDIA GTX-Titan GPU, and 64GB RAM.

Without a lexicon, the model takes less than 2ms recognizing an image. With a lexicon, recognition speed depends on the lexicon size. We adopt the precise search (Sec.~\ref{sec:recog-lexicon}) when lexicon size $\leq$ 1k. On larger lexicons, we adopt the approximate beam search (Sec.~\ref{sec:recog-lexicon}) with a beam width of 7. With a 50k-word lexicon, the search takes $\sim$200ms per image.

\subsection{Results on General Benchmarks}

Our model is firstly evaluated on benchmarks that are designed for general scene text recognition tasks. Samples in these benchmarks mostly contain regular text, but irregular text also exists. The benchmark datasets are:
\begin{itemize}
\item \textbf{IIIT 5K-Words}~\cite{MishraAJ12} (IIIT5K) contains 3000 cropped word images for testing. The images are collected from the Internet. For each image, there is a 50-word lexicon and a 1000-word lexicon. All lexicons consist of a ground truth word and some randomly picked words.
\item \textbf{Street View Text}~\cite{WangBB11} (SVT) is collected from Google Street View. Its test dataset consists of 647 word images. Many images in SVT are severely corrupted by noise and blur, or have very low resolutions. Each sample is associated with a 50-word lexicon.
\item \textbf{ICDAR 2003}~\cite{LucasPSTWYANOYMZOWJTWL05} (IC03) contains 860 cropped word images, each associated with a 50-word lexicon defined by Wang \emph{et al.}~\cite{WangBB11}. Following~\cite{WangBB11}, we discard images that contain non-alphanumeric characters or have less than three characters. Besides, there is a ``full lexicon'' which contains all lexicon words, and the Hunspell~\cite{Hunspell} lexicon which has 50k words.
\item \textbf{ICDAR 2013}~\cite{KaratzasSUIBMMMAH13} (IC13) inherits most of its samples from IC03. After filtering  samples as done in IC03, the dataset contains 857 samples.
\end{itemize}

In Tab.~\ref{tbl:general-datasets} we report our results, and compare them with other methods. On unconstrained recognition tasks (recognizing without a lexicon), our model outperforms all the other methods in comparison. On IIIT5K, RARE outperforms prior art CRNN~\cite{ShiBY15} by nearly 4 percentages, indicating a clear improvement in performance. We observe that IIIT5K contains a lot of irregular text, especially curved text, while RARE has an advantage in dealing with irregular text. Note that, although our model falls behind \cite{JaderbergSVZ14a} on some datasets, our model differs from~\cite{JaderbergSVZ14a} in that it is able recognize random strings such as telephone numbers, while \cite{JaderbergSVZ14a} only recognizes words that are in its 90k-dictionary. On constrained recognition tasks (recognizing with a lexicon), RARE achieves state-of-the-art or highly competitive accuracies. On IIIT5K, SVT and IC03, constrained recognition accuracies are on par with \cite{JaderbergSVZ14a}, and slightly lower than~\cite{ShiBY15}.

We also train and test a model that contains only the SRN. As reported in the last row of Tab.~\ref{tbl:general-datasets}, we see that the SRN-only model is also a very competitive recognizer, achieving higher or competitive performance on most of the benchmarks.

\begin{figure}[ht]
  \begin{centering}
  \includegraphics[width=0.95\linewidth]{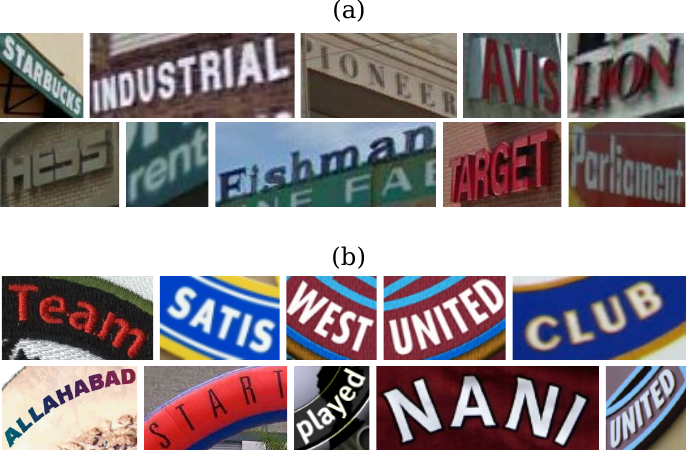}
  \par\end{centering}
  \vspace{0.2cm}
  \caption{Examples of irregular text. a) Perspective text. Samples are taken from the SVT-Perspective~\cite{PhanSTT13} dataset; b) Curved text. Samples are taken from the CUTE80~\cite{RisnumawanSCT14} dataset.}
  \label{fig:irregular-example}
\end{figure}

\subsection{Recognizing Perspective Text}

To validate the effectiveness of the rectification scheme, we evaluate RARE on the task of perspective text recognition. \textbf{SVT-Perspective}~\cite{PhanSTT13} is specifically designed for evaluating performance of perspective text recognition algorithms. Text samples in SVT-Perspective are picked from side view angles in Google Street View, thus most of them are heavily deformed by perspective distortion. Some examples are shown in Fig.~\ref{fig:irregular-example}.a. SVT-Perspective consists of 639 cropped images for testing. Each image is associated with a 50-word lexicon, which is inherited from the SVT~\cite{WangBB11} dataset. In addition, there is a ``Full'' lexicon which contains all the per-image lexicon words.

We use the same model trained on the synthetic dataset without fine-tuning. For comparison, we test the CRNN model~\cite{ShiBY15} on SVT-Perspective. We also compare RARE with \cite{WangBB11, MishraAJ12, WangWCN12, PhanSTT13}, whose recognition accuracies are reported in~\cite{PhanSTT13}.

\begin{table}[ht]
  \caption{Recognition accuracies on SVT-Perspective~\cite{PhanSTT13}. ``50'' and ``Full'' represent recognition with 50-word lexicons and the full lexicon respectively. ``None'' represents recognition without a lexicon.}
  \vspace{0.2cm}
  \footnotesize
  \begin{centering}
  \begin{tabular}{lc>{\centering}p{1.2cm}c}
    \toprule
    \textbf{Method} & \textbf{50} & \textbf{Full} & \textbf{None}\\
    \midrule
    Wang \emph{et al.} \cite{WangBB11} & 40.5 & 26.1 & -\\
    Mishra \emph{et al.} \cite{MishraAJ12} & 45.7 & 24.7 & -\\
    Wang \emph{et al.} \cite{WangWCN12} & 40.2 & 32.4 & -\\
    Phan \emph{et al.} \cite{PhanSTT13} & 75.6 & 67.0 & -\\
    Shi \emph{et al.} \cite{ShiBY15} & \textbf{92.6} & 72.6 & 66.8\\
    \textbf{RARE} & 91.2 & \textbf{77.4} & \textbf{71.8}\\
    \bottomrule
  \end{tabular}
  \par\end{centering}
  \label{tbl:svt-perspective}
\end{table}

\begin{figure}[htb]
  \begin{centering}
  \includegraphics[width=1\linewidth]{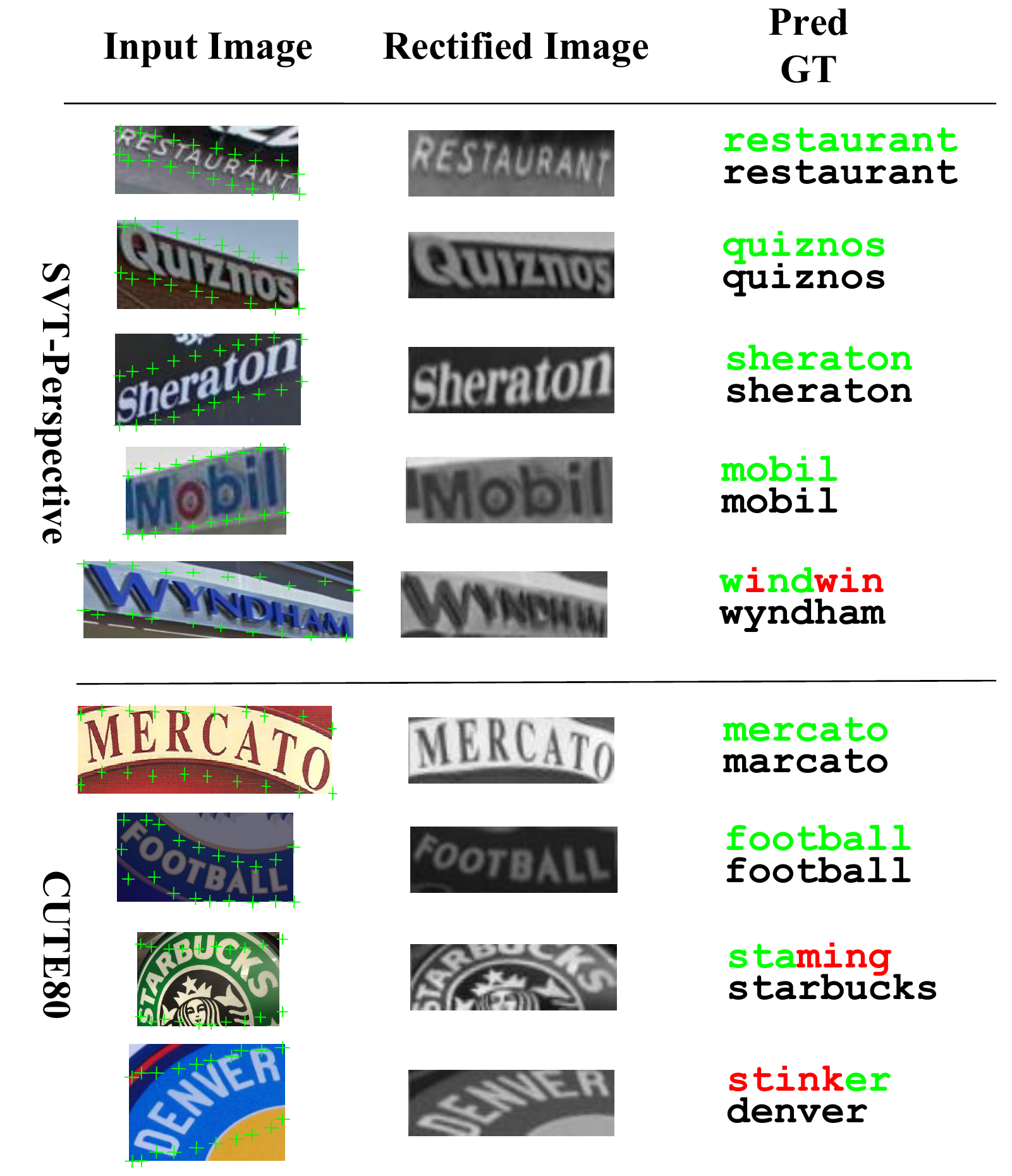}
  \par\end{centering}
  \vspace{0.2cm}
  \caption{Examples showing the rectifications our model makes and the recognition results. The left column is the input images, where green crosses are the predicted fiducial points. The middle column is the rectified images (we use gray-scale images for recognition). The right column is the recognized text and the ground truth text. Green and red characters are correctly and mistakenly recognized characters, respectively. The first five rows are taken from SVT-Perspective~\cite{PhanSTT13}, the rest rows are taken from CUTE80~\cite{RisnumawanSCT14}.}
  \label{fig:rectification-show}
\end{figure}

Tab.~\ref{tbl:svt-perspective} summarizes the results. In the second and third columns, we compare the accuracies of recognition with the 50-word lexicon and the full lexicon. Our method outperforms~\cite{PhanSTT13}, which is a perspective text recognition method, by a large margin on both lexicons. However, this gap is partially due to that we use a much larger training set than~\cite{PhanSTT13}. In the comparisons with~\cite{ShiBY15}, which uses the same training set as RARE, we still observe significant improvements in both the Full lexicon and the lexicon-free settings. Furthermore, recall the results in Tab.~\ref{tbl:general-datasets}, on SVT-Perspective RARE outperforms \cite{ShiBY15} by a even larger margin. The reason is that the SVT-perspective dataset mainly consists of perspective text, which is inappropriate for direct recognition. Our rectification scheme can significantly alleviate this problem.

In Fig.~\ref{fig:rectification-show} we present some qualitative analysis. Fiducial points predicted by the STN are plotted on input images in green crosses. We see that the STN tends to place fiducial points along upper and lower edges of scene text, and hence produces rectified images that are more readable for the SRN. However, the STN fails sometimes in the case of heavy perspective distortion.

\subsection{Recognizing Curved Text}

Curved text is a commonly seen artistic-style text in natural scenes. Due to its irregular character placement, recognizing curved text is very challenging. \textbf{CUTE80}~\cite{RisnumawanSCT14} focuses on the recognition of curved text. The dataset contains 80 high-resolution images taken in natural scenes. Originally, the dataset is proposed for detection tasks. We crop the words, resulting in 288 word images for testing. For comparisons, we evaluate the trained models of \cite{JaderbergSVZ14a} and \cite{ShiBY15}. All models are evaluated without a lexicon.

\begin{table}[hbt]
  \caption{Recognition accuracies on CUTE80~\cite{PhanSTT13}.}
  \vspace{0.2cm}
  \begin{centering}
  \footnotesize
  \begin{tabular}{lc}
    \toprule
    \addlinespace
    \textbf{Method} & \textbf{Accuracy}\\
    \midrule
    \addlinespace
    Jaderberg \emph{et al.} \cite{JaderbergSVZ14a} & 42.7\\
    \addlinespace
    Shi \emph{et al.} \cite{ShiBY15} & 54.9\\
    \addlinespace
    \textbf{RARE} & \textbf{59.2}\\
    \bottomrule
  \end{tabular}
  \par\end{centering}
  \label{tbl:cute80}
\end{table}

From the results summarized in Tab.~\ref{tbl:cute80}, we see that RARE outperforms the other two methods by a large margin. \cite{JaderbergSVZ14a} is a constrained recognition model, it cannot recognize words that are not in its dictionary. \cite{ShiBY15} is able to recognize arbitrary words, but it does not have a specific mechanism for handling curved text. Our model rectifies images that contain curved text before recognizing them. Therefore, it is advantageous on this task.

In Fig.~\ref{fig:rectification-show}, we demonstrate the effect of rectification through some examples. Generally, the rectification made by the STN is not perfect, but it alleviates the recognition difficulty to some extent. RARE tends to fail when curve angles are too large, as shown in the last two rows of Fig.~\ref{fig:rectification-show}.

\section{Conclusion}

We study a common but difficult problem in scene text recognition, called the irregular text problem. Traditional solutions typically use a separate text rectification component. We address this problem in a more feasible and elegant way by adopting a differentiable spatial transformer network module. In addition, the spatial transformer network is connected to an attention-based sequence recognizer, allowing us to train the whole model end-to-end. The extensive experimental results show that 1) without geometric supervision, the learned model can automatically generate more ``readable'' images for both human and the sequence recognition network; 2) the proposed text rectification method can significantly improve recognition accuracies on irregular scene text; 3) the proposed scene text recognition system is competitive compared with the state-of-the-arts. In the future, we plan to address the end-to-end scene text reading problem through the combination of RARE with a scene text detection method, \emph{e.g.}~\cite{ZhangZSLB16}.

\section*{Acknowledgments}

% \paragraph{Acknowledgments}
This work was primarily supported by National Natural Science Foundation of China (NSFC) (No. 61222308, No. 61573160 and No. 61503145), and Open Project Program of the State Key Laboratory of Digital Publishing Technology (No. F2016001).

{\small
\bibliographystyle{ieee}
\bibliography{references}

\begin{thebibliography}{10}\itemsep=-1pt

\bibitem{Hunspell}
Hunspell.
\newblock \url{http://hunspell.sourceforge.net/}.

\bibitem{AlmazanGFV14}
J.~Almaz{\'{a}}n, A.~Gordo, A.~Forn{\'{e}}s, and E.~Valveny.
\newblock Word spotting and recognition with embedded attributes.
\newblock {\em IEEE Trans. Pattern Anal. Mach. Intell.}, 36(12):2552--2566,
  2014.

\bibitem{AlsharifP13}
O.~Alsharif and J.~Pineau.
\newblock End-to-end text recognition with hybrid hmm maxout models.
\newblock {\em ICLR}, 2014.

\bibitem{BahdanauCB14}
D.~Bahdanau, K.~Cho, and Y.~Bengio.
\newblock Neural machine translation by jointly learning to align and
  translate.
\newblock {\em CoRR}, abs/1409.0473, 2014.

\bibitem{BissaccoCNN13}
A.~Bissacco, M.~Cummins, Y.~Netzer, and H.~Neven.
\newblock Photoocr: Reading text in uncontrolled conditions.
\newblock In {\em ICCV}, 2013.

\bibitem{Bookstein89}
F.~L. Bookstein.
\newblock Principal warps: Thin-plate splines and the decomposition of
  deformations.
\newblock {\em IEEE Trans. Pattern Anal. Mach. Intell.}, 11(6):567--585, 1989.

\bibitem{ChoMBB14}
K.~Cho, B.~van Merrienboer, D.~Bahdanau, and Y.~Bengio.
\newblock On the properties of neural machine translation: Encoder-decoder
  approaches.
\newblock {\em CoRR}, abs/1409.1259, 2014.

\bibitem{ChorowskiBSCB15}
J.~Chorowski, D.~Bahdanau, D.~Serdyuk, K.~Cho, and Y.~Bengio.
\newblock Attention-based models for speech recognition.
\newblock {\em CoRR}, abs/1506.07503, 2015.

\bibitem{Collobert11}
R.~Collobert, K.~Kavukcuoglu, and C.~Farabet.
\newblock Torch7: A matlab-like environment for machine learning.
\newblock In {\em BigLearn, NIPS Workshop}, 2011.

\bibitem{DalalT05}
N.~Dalal and B.~Triggs.
\newblock Histograms of oriented gradients for human detection.
\newblock In {\em CVPR}, 2005.

\bibitem{GoelMAJ13}
V.~Goel, A.~Mishra, K.~Alahari, and C.~V. Jawahar.
\newblock Whole is greater than sum of parts: Recognizing scene text words.
\newblock In {\em ICDAR}, 2013.

\bibitem{Gordo14}
A.~Gordo.
\newblock Supervised mid-level features for word image representation.
\newblock In {\em CVPR}, 2015.

\bibitem{GravesMH13}
A.~Graves, A.~Mohamed, and G.~E. Hinton.
\newblock Speech recognition with deep recurrent neural networks.
\newblock In {\em ICASSP}, 2013.

\bibitem{HochreiterS97}
S.~Hochreiter and J.~Schmidhuber.
\newblock Long short-term memory.
\newblock {\em Neural Computation}, 9(8):1735--1780, 1997.

\bibitem{JaderbergSVZ14}
M.~Jaderberg, K.~Simonyan, A.~Vedaldi, and A.~Zisserman.
\newblock Synthetic data and artificial neural networks for natural scene text
  recognition.
\newblock {\em NIPS Deep Learning Workshop}, 2014.

\bibitem{JaderbergSVZ14b}
M.~Jaderberg, K.~Simonyan, A.~Vedaldi, and A.~Zisserman.
\newblock Deep structured output learning for unconstrained text recognition.
\newblock In {\em ICLR}, 2015.

\bibitem{JaderbergSVZ14a}
M.~Jaderberg, K.~Simonyan, A.~Vedaldi, and A.~Zisserman.
\newblock Reading text in the wild with convolutional neural networks.
\newblock {\em Int. J. Comput. Vision}, 2015.

\bibitem{JaderbergSZK15}
M.~Jaderberg, K.~Simonyan, A.~Zisserman, and K.~Kavukcuoglu.
\newblock Spatial transformer networks.
\newblock {\em CoRR}, abs/1506.02025, 2015.

\bibitem{JaderbergVZ14}
M.~Jaderberg, A.~Vedaldi, and A.~Zisserman.
\newblock Deep features for text spotting.
\newblock In {\em ECCV}, 2014.

\bibitem{KaratzasSUIBMMMAH13}
D.~Karatzas, F.~Shafait, S.~Uchida, M.~Iwamura, L.~G. i~Bigorda, S.~R. Mestre,
  J.~Mas, D.~F. Mota, J.~Almaz{\'{a}}n, and L.~de~las Heras.
\newblock {ICDAR} 2013 robust reading competition.
\newblock In {\em ICDAR}, 2013.

\bibitem{KrizhevskySH12}
A.~Krizhevsky, I.~Sutskever, and G.~E. Hinton.
\newblock Imagenet classification with deep convolutional neural networks.
\newblock In {\em NIPS}, 2012.

\bibitem{LecunLYP1998}
Y.~LeCun, L.~Bottou, Y.~Bengio, and P.~Haffner.
\newblock Gradient-based learning applied to document recognition.
\newblock {\em Proceedings of the IEEE}, 86(11):2278--2324, 1998.

\bibitem{Lowe04}
D.~G. Lowe.
\newblock Distinctive image features from scale-invariant keypoints.
\newblock {\em Int. J. Comput. Vision}, 60(2):91--110, 2004.

\bibitem{LucasPSTWYANOYMZOWJTWL05}
S.~M. Lucas, A.~Panaretos, L.~Sosa, A.~Tang, S.~Wong, R.~Young, K.~Ashida,
  H.~Nagai, M.~Okamoto, H.~Yamamoto, H.~Miyao, J.~Zhu, W.~Ou, C.~Wolf,
  J.~Jolion, L.~Todoran, M.~Worring, and X.~Lin.
\newblock {ICDAR} 2003 robust reading competitions: entries, results, and
  future directions.
\newblock {\em IJDAR}, 7(2-3):105--122, 2005.

\bibitem{MishraAJ12}
A.~Mishra, K.~Alahari, and C.~V. Jawahar.
\newblock Scene text recognition using higher order language priors.
\newblock In {\em BMVC}, 2012.

\bibitem{Nagy00}
G.~Nagy.
\newblock Twenty years of document image analysis in {PAMI}.
\newblock {\em IEEE Trans. Pattern Anal. Mach. Intell.}, 22(1):38--62, 2000.

\bibitem{NairH10}
V.~Nair and G.~E. Hinton.
\newblock Rectified linear units improve restricted boltzmann machines.
\newblock In {\em ICML}, 2010.

\bibitem{NeumannM12}
L.~Neumann and J.~Matas.
\newblock Real-time scene text localization and recognition.
\newblock In {\em CVPR}, 2012.

\bibitem{PhanSTT13}
T.~Q. Phan, P.~Shivakumara, S.~Tian, and C.~L. Tan.
\newblock Recognizing text with perspective distortion in natural scenes.
\newblock In {\em ICCV}, 2013.

\bibitem{RisnumawanSCT14}
A.~Risnumawan, P.~Shivakumara, C.~S. Chan, and C.~L. Tan.
\newblock A robust arbitrary text detection system for natural scene images.
\newblock {\em Expert Syst. Appl.}, 41(18):8027--8048, 2014.

\bibitem{Rodriguez-Serrano15}
J.~A. Rodr{\'{\i}}guez{-}Serrano, A.~Gordo, and F.~Perronnin.
\newblock Label embedding: {A} frugal baseline for text recognition.
\newblock {\em Int. J. Comput. Vision}, 113(3):193--207, 2015.

\bibitem{ShiBY15}
B.~Shi, X.~Bai, and C.~Yao.
\newblock An end-to-end trainable neural network for image-based sequence
  recognition and its application to scene text recognition.
\newblock {\em CoRR}, abs/1507.05717, 2015.

\bibitem{SimonyanZ14a}
K.~Simonyan and A.~Zisserman.
\newblock Very deep convolutional networks for large-scale image recognition.
\newblock {\em CoRR}, abs/1409.1556, 2014.

\bibitem{SuL14}
B.~Su and S.~Lu.
\newblock Accurate scene text recognition based on recurrent neural network.
\newblock In {\em ACCV}, 2014.

\bibitem{WangBB11}
K.~Wang, B.~Babenko, and S.~Belongie.
\newblock End-to-end scene text recognition.
\newblock In {\em ICCV}, 2011.

\bibitem{WangB10}
K.~Wang and S.~Belongie.
\newblock Word spotting in the wild.
\newblock In {\em ECCV}, 2010.

\bibitem{WangWCN12}
T.~Wang, D.~J. Wu, A.~Coates, and A.~Y. Ng.
\newblock End-to-end text recognition with convolutional neural networks.
\newblock In {\em ICPR}, 2012.

\bibitem{YaoBLMT12}
C.~Yao, X.~Bai, W.~Liu, Y.~Ma, and Z.~Tu.
\newblock Detecting texts of arbitrary orientations in natural images.
\newblock In {\em CVPR}, 2012.

\bibitem{YaoBSL14}
C.~Yao, X.~Bai, B.~Shi, and W.~Liu.
\newblock Strokelets: {A} learned multi-scale representation for scene text
  recognition.
\newblock In {\em CVPR}, 2014.

\bibitem{YeD15}
Q.~Ye and D.~S. Doermann.
\newblock Text detection and recognition in imagery: {A} survey.
\newblock {\em IEEE Trans. Pattern Anal. Mach. Intell.}, 37(7):1480--1500,
  2015.

\bibitem{Matthew12ADADELTA}
M.~D. Zeiler.
\newblock {ADADELTA:} an adaptive learning rate method.
\newblock {\em CoRR}, abs/1212.5701, 2012.

\bibitem{ZhangGLM12}
Z.~Zhang, A.~Ganesh, X.~Liang, and Y.~Ma.
\newblock {TILT:} transform invariant low-rank textures.
\newblock {\em Int. J. Comput. Vision}, 99(1):1--24, 2012.

\bibitem{ZhangZSLB16}
Z.~Zhang, C.~Zhang, W.~Shen, C.~Yao, W.~Liu, and X.~Bai.
\newblock Multi-oriented text detection with fully convolutional networks.
\newblock In {\em CVPR}, 2016.

\bibitem{ZhuYB16}
Y.~Zhu, C.~Yao, and X.~Bai.
\newblock Scene text detection and recognition: recent advances and future
  trends.
\newblock {\em Frontiers of Computer Science}, 10(1):19--36, 2016.

\end{thebibliography}
}

\end{document}